\documentclass[final]{cvpr}

\usepackage{comment}
\usepackage{caption}
\usepackage{stfloats}
\usepackage{times}
\usepackage{epsfig}
\usepackage{graphicx}
\usepackage{amsmath}
\usepackage{amssymb}
\usepackage{ctable}
\usepackage{colortbl}
\usepackage{lipsum}
\usepackage{xcolor}

\usepackage{multirow}

\usepackage{subcaption}

\usepackage[pagebackref=true,breaklinks=true,colorlinks,bookmarks=false]{hyperref}

\def\cvprPaperID{****} 
\def\confYear{CVPR 2021}

\begin{document}
\pagestyle{empty} 

\newcommand{\OURS}{Contrastive Scene Contexts}
\newcommand{\OURSSHORT}{CSC}
\newcommand{\JI}[1]{\textbf{\textcolor{blue}{JI: #1}}}
\newcommand{\BEN}[1]{\textbf{\textcolor{green}{BEN: #1}}}
\newcommand{\MATTHIAS}[1]{\textbf{\textcolor{purple}{MATTHIAS: #1}}}
\newcommand{\SAINING}[1]{\textbf{\textcolor{red}{SAINING: #1}}}
\definecolor{Gray}{gray}{0.92}
\definecolor{darkgreen}{rgb}{0.13, 0.55, 0.13}

\title{Exploring Data-Efficient 3D Scene Understanding \\ with Contrastive Scene Contexts}
\author{%
Ji Hou$^{1}$~~~~Benjamin Graham$^{2}$~~~~Matthias Nie{\ss}ner$^{1}$~~~~Saining Xie$^{2}$ \vspace{0.2cm}\\
$^{1}$Technical University of Munich~~~~$^{2}$Facebook AI Research
}
\twocolumn[{%
\renewcommand\twocolumn[1][]{#1}%
\maketitle
\thispagestyle{empty}

\begin{center}
\vspace{-0.4cm}
\includegraphics[width=0.85\linewidth]{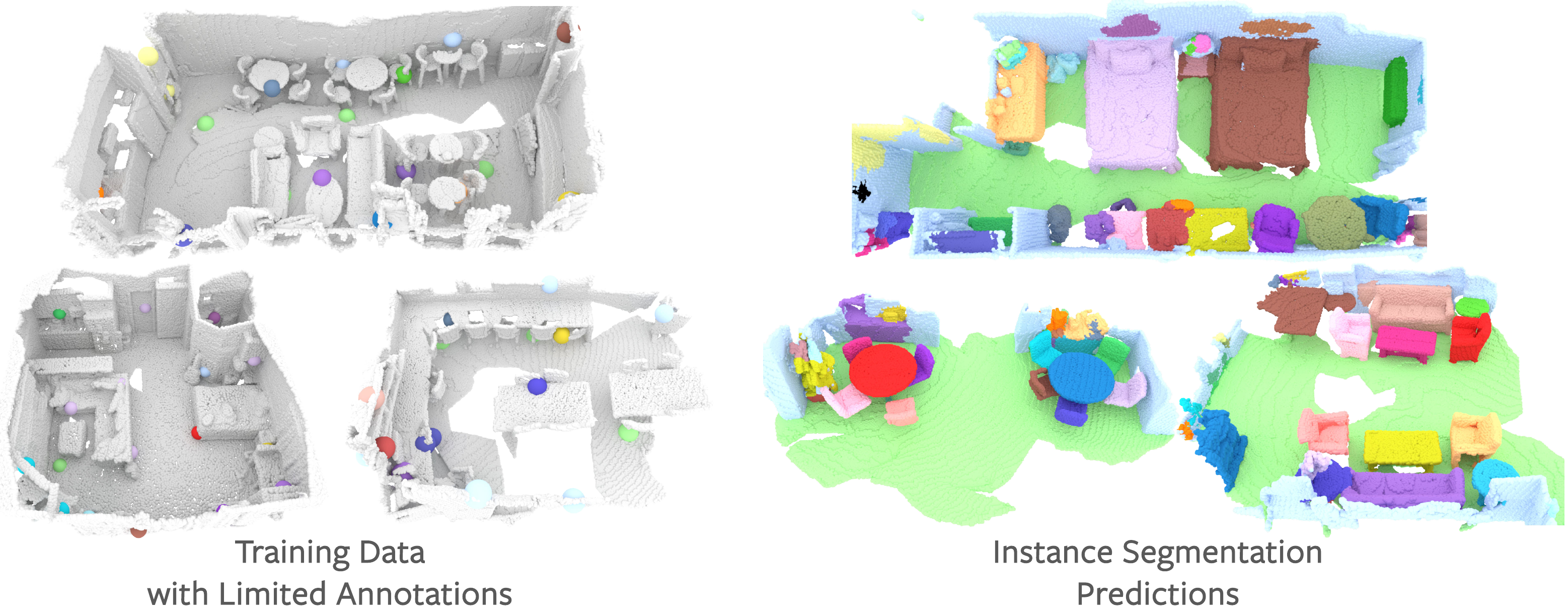}
\captionof{figure}{
How many point labels are necessary to train a 3D instance segmentation model on point clouds? It turns out not too many! With the help of unsupervised pre-training, only 20 labelled points per scene (\emph{less than 0.1\% of the total points}) are used to fine-tune an instance segmentation model on ScanNet. \textbf{Left}: Train samples; only colored points (enlarged for better visibility) are labeled. \textbf{Right}: Predictions in validation set and different colors represent different instances.}
\label{fig:teaser}
\end{center}
}]

\begin{abstract}
\vspace{-0.5cm}
The rapid progress in 3D scene understanding has come with growing demand for data; however, collecting and annotating 3D scenes (\eg point clouds) are notoriously hard. For example, the number of scenes (\eg indoor rooms) that can be accessed and scanned might be limited; even given sufficient data, acquiring 3D labels (\eg instance masks) requires intensive human labor. In this paper, we explore data-efficient learning for 3D point cloud. As a first step towards this direction, we propose Contrastive Scene Contexts, a 3D pre-training method that makes use of both point-level correspondences and spatial contexts in a scene. Our method achieves state-of-the-art results on a suite of benchmarks where training data or labels are scarce. Our study reveals that exhaustive labelling of 3D point clouds might be unnecessary; and remarkably, on ScanNet, even using 0.1\% of point labels, we still achieve 89\% (instance segmentation) and 96\% (semantic segmentation) of the baseline performance that uses full annotations~\footnote{Code is available at \href{https://github.com/facebookresearch/ContrastiveSceneContexts}{GitHub}}.

\end{abstract}

\section{Introduction}
\vspace{-0.5em}
Recent advances in deep learning on point clouds, such as those obtained from LiDAR or depth sensors, together with a proliferation of public, annotated datasets \cite{shapenet2015, dai2017scannet, silberman2012indoor, janoch2013category, xiao2013sun3d, armeni_cvpr16, Mo_2019_CVPR, Sun_2020_CVPR}, have led to swift progress in 3D scene understanding. However, compared to large-scale 2D scene understanding on images \cite{deng2009imagenet, lin2014microsoft, gupta2019lvis}, the scale of 3D scene understanding---in terms of the amount and diversity of data and annotations, the model size, the number of semantic categories, and so on---still falls behind. We argue that one major bottleneck is the fact that collecting and annotating diverse 3D scenes are significantly more expensive. Unlike 2D images that comfortably exists on the Internet, collecting real world 3D scene datasets usually involves traversing the environment in real life and scanning with 3D sensors. Therefore, the number of indoor scenes that can be scanned might be limited.
What is more concerning is that, even given sufficient data acquisition, 3D semantic labelling (\eg bounding boxes and instance masks) requires complex pipelines \cite{dai2017scannet} and labor-intensive human effort.

In this work, we explore a new learning task in 3D, \ie data-efficient 3D scene understanding, which focuses on the problem of learning with limited data or supervision\footnote{Sometimes a distinction is drawn between \emph{data-efficiency} and \emph{label-efficiency}, to separate the scenarios of limited amount of data samples and limited supervision; here, we use \emph{data-efficiency} to encompass both cases.}. We note that the importance of data-efficient learning in 3D is two-fold. One concerns the status quo: given limited data we have right now, \emph{can we design better methods that perform better?} The other one is more forward-looking: \emph{is it possible to reduce the human labor for annotation}, with a goal of creating 3D scene datasets on a much larger scale?

To formally study the problem, we first introduce a suite of scene understanding benchmarks that encompasses two complementary settings for data-efficient learning: (1) \emph{limited scene reconstructions (LR)} and (2) \emph{limited annotations (LA)}. The first setting concerns the scenario where the bottleneck is the number of scenes that can be scanned and reconstructed. The second one focuses on the case where in each scene, the budget for labeling is constrained (\eg one can only label a small set of points). For each setting, the evaluation is done on a diverse set of scene understanding tasks including object detection, semantic segmentation and instance segmentation.

For data-efficient learning in 2D \cite{henaff2019data}, representation learning, \eg pre-training on a rich source set and fine-tuning on a much smaller target set, often comes to the rescue; in 3D, representation learning for data-efficient learning is even more wanted but long overdue. With this perspective, we focus on studying data-efficient 3D scene understanding through the lens of representation learning. 

Only recently, PointContrast~\cite{xie2020pointcontrast} demonstrates that network weights pre-trained on 3D partial frames can lead to a performance boost when fine-tuned on 3D semantic segmentation and object detection tasks. Our work is inspired by PointContrast. However, we observe that the simple contrastive-learning based pretext task used in~\cite{xie2020pointcontrast} only concerns point-level correspondence matching, which completely disregards the spatial configurations and contexts in a scene. In Section~\ref{method}, we show that this design limits the scalibility and transferability; we further propose an approach that integrates the spatial information into the contrastive learning framework. The simple modification can significantly improve the performance over PointContrast, especially on complex tasks such as instance segmentation.

Our exploration in data-efficient 3D scene understanding provides some surprising observations. For example, on ScanNet, even using 0.1\% of point labels, we are still able to recover 89\% (instance segmentation) and 96\% (semantic segmentation) of the baseline performance that uses full annotations. The results imply that exhaustive labelling of 3D point clouds might not be necessary. In both scenarios of \emph{limited scene reconstructions (LR)} and \emph{limited annotations (LA)}, our pre-trained network, when used as the initialization for supervised fine-tuning, offers consistent improvement across multiple tasks and datasets. In the scenario of \emph{LA}, we also show that an active labeling strategy can be enabled by clustering the pre-trained point features.

In summary, the contributions of our work include:
\begin{itemize} \itemsep0em 
    \item A systematic study on data-efficient 3D scene understanding with a comprehensive suite of benchmarks.
    \item A new 3D pre-training method that can gracefully transfer to complex tasks such as instance segmentation and outperform the state-of-the-art results. 
    \item Given the pre-trained network, we study practical solutions for data-efficient learning in 3D through fine-tuning as well as an active labeling strategy.
\end{itemize}
\section{Related Work}
\noindent \textbf{3D Scene Understanding.} Research in deep learning on 3D point clouds have been recently shifted from synthetic, single object classification~\cite{qi2016volumetric,qi2017pointnet, qi2017pointnetplusplus} to the challenge of large-scale, real-world scene understanding. A variety of datasets~\cite{armeni_cvpr16, dai2017scannet, song2015sun, geiger2012we, Sun_2020_CVPR} and algorithms have been proposed for 3D object detection~\cite{qi2018frustum, voteNet, imvotenet, gwak2020generative}, semantic segmentation~\cite{qi2017pointnet, tchapmi2017segcloud,wu2019pointconv,thomas2019kpconv,graham20183d, choy20194d} and instance segmentation~\cite{wang2018sgpn, hou20193d, yi2018gspn,lahoud20193d, wang2019associatively, yang2019learning, hou2020revealnet, engelmann20203d, jiang2020pointgroup, jiang2020end}. In the past year, sparse convolutional networks~\cite{graham20183d, choy20194d} stand out as a promising approach to standardize deep learning for point clouds, due to its computational efficiency and state-of-the-art performance for 3D scene understanding tasks~\cite{choy20194d, han2020occuseg, jiang2020pointgroup}. In this work, we also adopt a sparse U-Net~\cite{ronneberger2015u} backbone for our exploration.

\noindent \textbf{3D Representation Learning.} Compared to 2D vision, the limits of big data are far from being fully explored in 3D. In 2D representation learning, for example, transfer learning from a rich source data (\eg ImageNet~\cite{deng2009imagenet}) to a (typically smaller) target data, has become a dominant framework for many applications~\cite{girshick2014rich}. In contrast, 3D representation learning has not been widely adopted and most 3D networks are trained from scratch on the target data directly. Recently, unsupervised pre-training has made great progress and drawn significant attention in 2D~ \cite{oord2018representation,bachman2019learning,misra2019self,henaff2019data,wu2018unsupervised,tian2019contrastive, hjelm2018learning,he2020momentum,chen2020simple,caron2020unsupervised,grill2020bootstrap}. Following suit, recent works attempt to adapt the 2D pretext tasks to 3D, but mostly focus on single object classification tasks on ShapeNet \cite{achlioptas2017learning,gadelha2018multiresolution,yang2018foldingnet,groueix2018papier,li2018so,wang2019deep,hassani2019unsupervised,sauder2019self,sanghi2020info3d}. Our work is mostly inspired by a recent contrastive-learning based method PointContrast~\cite{xie2020pointcontrast}, which first demonstrates the effectiveness of unsupervised pre-training on a diverse set of scene-level understanding tasks. As we will show in the later sections, the simple point-level pre-training objective in PointContrast ignores the spatial contexts of the scene (such as relative poses of objects, and distances between them) which limits its transferability for complex tasks such as instance segmentation. PointContrast also focuses on downstream tasks with 100\% data and labels, while we systematically explore a new data-efficient paradigm that has practical importance.

\noindent \textbf{Data-Efficient Learning.} 
Data-efficient learning concerns the problem of learning with limited training examples or labels. This capability is known in cognitive science to be a distinctive characteristic of humans~\cite{biederman1987recognition}. In contrast, training deep neural networks is not naturally data-efficient, as it typically relies on large amount of annotated data. Among many potential solutions towards this goal, representation learning (commonly through transfer learning) is arguably the most promising one. A good representation ``\emph{entangles the different explanatory factors of variation behind the data}''~\cite{bengio2013representation} and thus makes the downstream prediction easier (and less data-hungry). This concept has been validated successfully in natural language processing~\cite{brown2020language} and to some extent in 2D image classification~\cite{henaff2019data}. Pursuing this direction in 3D is even more desirable, considering the potential benefit in reducing the labor of data collection and annotation. Existing work focuses on mostly single CAD model classification or part segmentation~\cite{zhu2019cosegnet,sharma2019learning,muralikrishnan2018tags2parts,chen2019bae,hassani2019unsupervised,gadelha2020label,xu2020weakly}. To the best of our knowledge, our work is the first to systematically explore data-efficient learning in a real-world, large-scale 3D scene understanding (on semantic/instance segmentation and detection) setup.
\section{Contrastive Scene Contexts for Pre-training}
\label{method}
\vspace{-0.5em}

In this section, we first briefly revisit the PointContrast framework~\cite{xie2020pointcontrast}, and discuss the shortcomings and remedies. We then introduce our pre-training algorithm.
\vspace{0.2em}

\noindent \textbf{Revisiting PointContrast.} The pre-training objective for PointContrast is to achieve point \emph{equivariance} with respect to a set of random geometric transformations. Given a pair of overlapping partial scans, a contrastive loss for pre-training is defined over the point features. The objective is to minimizes the distance for matched points (positive pairs) and maximize the distance between unmatched ones (negative pairs). Despite the fact that strong spatial contexts exist among objects in a scene, this objective does not capture any of the spatial information: the negative pairs could be sampled from arbitrary locations across many scenes in a mini-batch. We hypothesize that this leads to some limitations: 1) the spatial contexts (\eg relative pose, direction and distance), which could be pivotal for complex tasks such as instance segmentation, are entirely discarded from pre-training; 2) the scalibility of contrastive learning might be hampered; PointContrast cannot utilize a large number of negative points, potentially because that contrasting a pair of spatially distant and unrelated points would contribute little to learning. In fact, PointContrast uses only a random sampling of 1024 points per scene for pre-training, and it has been shown that results do not improve with more sampled points~\cite{xie2020pointcontrast}. We also confirm this behavior with experiments later this section.

\begin{figure}[t!]
\begin{center}
\includegraphics[width=1.0\linewidth]{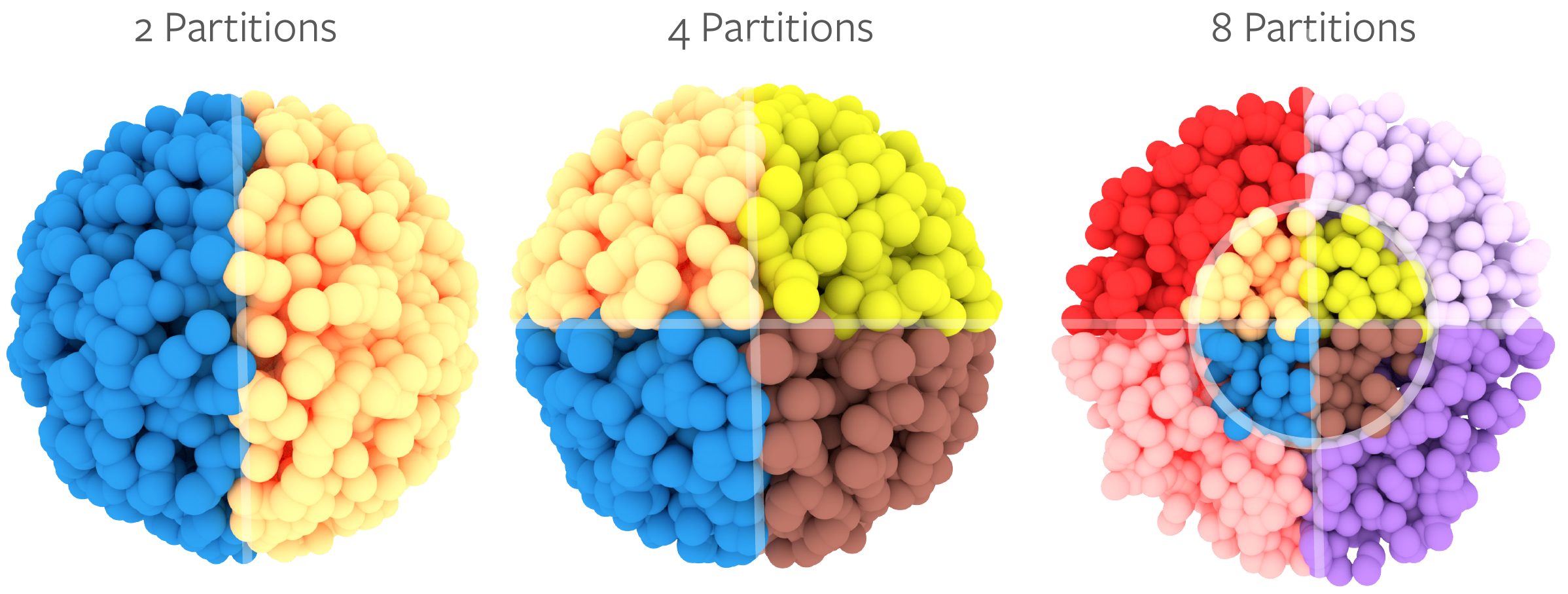}
\end{center}
\vspace{-0.4cm}
\caption{\textbf{Illustration of Scene Contexts.} We visualize the 2,4 and 8 spatial partitions for Scene Contexts. The anchor point is in the center. For 2 and 4 partitions, only relative angles are sufficient. For 8 partitions (a cross-section is shown), both relative angles and distances are needed.}
\label{fig:partitions}
\vspace{-0.3cm}
\end{figure}

\noindent \textbf{Contrastive Scene Contexts.} We hope to integrate spatial contexts into the pre-training objective. There are many ways to achieve the goal, and here we take inspiration from the classic \emph{ShapeContext} local descriptor~\cite{belongie2002shape, kortgen20033d, xie2018attentional} for shape matching. The \emph{ShapeContext} descriptor partitions the space into spatially inhomogeneous cells, and encodes the spatial contexts about the shape at each point by computing a histogram over the number of neighboring points in each cell. We call our method \emph{Contrastive Scene Contexts} because at a high level, our method also aims to capture \emph{the distribution over relative locations in a scene}. We partition the scene point cloud into multiple regions, and instead of having a single contrastive loss for the entire point set sampled in a mini-batch, we perform contrastive learning in each region separately, and aggregate the losses in the end.

Concretely, given a pair of partial frame point clouds $\mathbf{x}$ and $\mathbf{y}$ from the same scene, we have correspondence mapping $(i,j) \in M_{\mathbf{x}\mathbf{y}}$ available, where $i$ is the index of a point $\mathbf{x}_i \in \mathcal{R}^3$ in frame $\mathbf{x}$ and $j$ is the index of a matched point $\mathbf{y}_j \in \mathcal{R}^3$ in frame $\mathbf{y}$. Similar to PointContrast, we sample $N$ pairs of matched points as positives. However, in our method, for each anchor point $\mathbf{x}_i$, the space is divided into multiple partitions and other points are assigned to different partitions based on their relative angles and distances to $i$. 

The distance and angle information needed for scene context partition at anchor point $\mathbf{x}_i$ is as follows,
\begin{align}
    &\mathcal{D}_{ik} = \sqrt{\sum_{d=1}^{3}(\mathbf{x}_i^d-\mathbf{x}_k^d)^2} \\
    &\mathcal{A}_{ik} = arctan2(\mathcal{D}_{ik}) + 2\pi
\end{align}
where $\mathcal{D}$ is the relative distance matrix. $\mathcal{D}_{ik}$ stores the distance between point $i$ and point $k$ and $\mathcal{A}$ is the relative angle matrix, where $A_{ik}$ stores the relative angle between point $i$ and point $k$. In Equation (1) $d$ represents the 3D dimension. With ${\mathcal{D}}$ and ${\mathcal{A}}$, a \emph{ShapeContext}-like spatial partitioning function can be easily constructed on-the-fly. In Figure~\ref{fig:partitions}, we show a visual illustration of how the space partitioning works. Computing 2 or 4 partitions only requires cutting the space according to relative angles based on $\mathcal{A}$; while the 8 or more partitions also require the extent of the inner regions using ${\mathcal{D}}$. We always partition the space uniformly along the relative angles and distances. Note that the partitioning is \emph{relative to the anchor point} $i$.

Suppose there are $P$ partitions, we denote the spatial partition functions as $par_p(\cdot)$, where $p\in{\{1,\dots,P\}}$. Function $par_p(\cdot)$ takes the anchor point $i$ as input, and return a set of points as negatives. A PointInfoNCE loss $\mathcal{L}_p$ is independently computed for each partition:
{
\footnotesize
\begin{align}
    &\mathcal{L}_p = -\sum_{(i,j) \in M}\log\frac{\exp(\mathbf{f}^1_i\cdot\mathbf{f}^2_j/\tau)}{\sum_{(\cdot, k)\in M,k\in par_{p}(i)}\exp(\mathbf{f}^1_i\cdot\mathbf{f}^2_k/\tau)}
\end{align}
}%
Details of Equation (3) and other implementation details can be found in Appendix. The final loss is computed by aggregating all partitions $\mathcal{L} = \frac{1}{|P|}\sum_p \mathcal{L}_p$.

\begin{figure}[t]
\begin{center}
\includegraphics[width=1.0\linewidth]{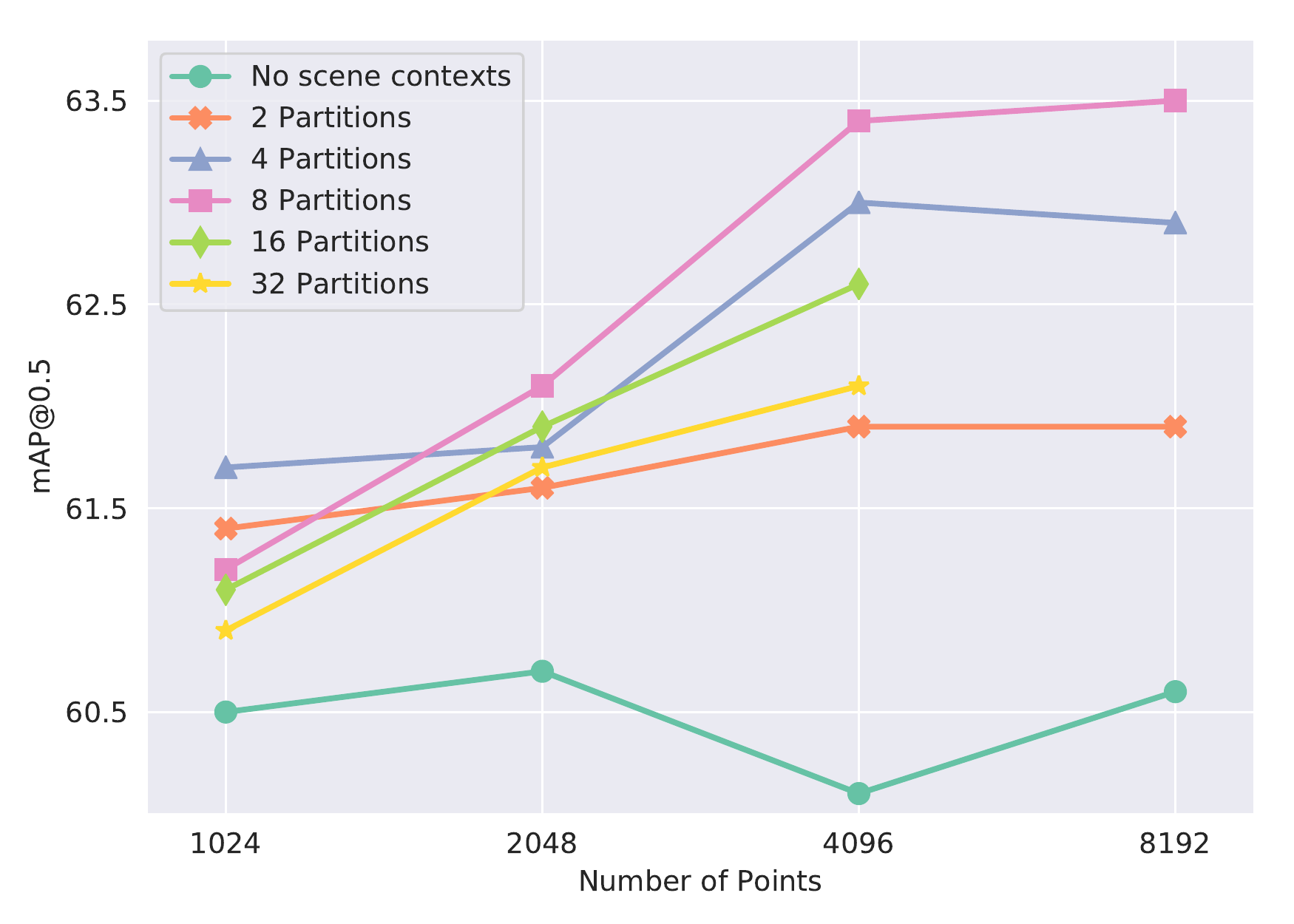}
\end{center}
\vspace{-20pt}
\caption{\textbf{Analysis Experiment.} Varying the number of partitions and sampled points for pre-training; Results are reported on the S3DIS instance segmentation task~\cite{armeni_cvpr16}. Using scene context partitions has enabled constrastive learning to utilize more points for better performance.}
\label{fig:ablation_shape_contexts}
\label{fig:shapecontext}
\end{figure}

\noindent \textbf{Analysis.} We first show that by integrating the scene contexts into the objective, our pre-training method can benefit more from a larger point set. We conduct an analysis experiment by varying the number of scene context partitions and the number of points sampled for computing the contrastive loss. We pre-train our model for a short schedule (20K iters). We then fine-tune the pre-trained weights on S3DIS instance segmentation benchmark \cite{armeni_cvpr16}. Results are shown in Figure~\ref{fig:ablation_shape_contexts}, the green line represents a variant with \emph{no spatial partitioning}; the left-most point represents PointContrast\footnote{Not exactly identical since the matched points are sampled per scene in this experiment, rather than from the whole mini-batch as in PointContrast; we have verified that this nuance does not influence the conclusion.}. Similar to the observation in \cite{xie2020pointcontrast}, without scene contexts, increasing the number of sampled points does not improve the performance; with more partitions, increasing \# sampled points leads to a consistent boost in performance (up to 4096 points). We use 8 partitions as empirically it works best. This shows that our method leads to better \emph{scalability} as more points can be utilized for pre-training.

We achieve state-of-the-art instance segmentation results in terms of mAP@0.5 (Table.~\ref{tab:ins_s3dis}) using a simple bottom-up clustering mechanism with voting loss (details in Appendix). We do not use any special modules such as Proposal Aggregation~\cite{engelmann20203d} or Scoring Network~\cite{jiang2020pointgroup}. We observe a $2.9\%$ absolute improvement over PointContrast pre-training, which brings the improvement over train-from-scratch baseline to $4.1\%$. This substantial margin demonstrates the effectiveness of Contrastive Scene Contexts on instance segmentation tasks. We provide more results comparing against PointContrast in Section~\ref{comp_to_pc}.

\begin{table}[h]
  \centering
  \small
  \begin{tabular}{c|l}
  \specialrule{1.1pt}{0.1pt}{0pt}
   Methods & mAP@0.5\\
  \hline
ASIS~\cite{wang2019associatively}    & \quad  55.3 \\
3D-BoNet~\cite{yang2019learning}     & \quad 57.5 \\
PointGroup~\cite{jiang2020pointgroup}& \quad 57.8  \\
3D-MPA~\cite{engelmann20203d}       &  \quad 63.1  \\
\hline
Train from scratch                   & \quad 59.3   \\
PointContrast (PointInfoNCE) ~\cite{xie2020pointcontrast}&  \quad 60.5~\tiny{\textcolor{gray}{(+1.2)}}  \\
\OURS & \quad \textbf{63.4}~\tiny{\textcolor{darkgreen}{(+4.1)}}  \\
 \specialrule{1.1pt}{0.1pt}{0pt}
  \end{tabular}
\vspace{-0.1cm}
\caption{\textbf{Fine-tuning results for instance segmentation on S3DIS~\cite{armeni_cvpr16}.} A simple clustering-based model with \emph{Contrastive Scene Contexts} pre-trained backbone performs significantly better than the train-from-scratch baseline and PointContrast pre-training~\cite{xie2020pointcontrast}.}.
\label{tab:ins_s3dis}
\vspace{-0.5cm}
\end{table}
\section{Data-Efficient 3D Scene Understanding}
\label{benchmark}
\vspace{-0.5em}

\begin{figure*}[h]
\begin{center}
\includegraphics[width=0.95\linewidth]{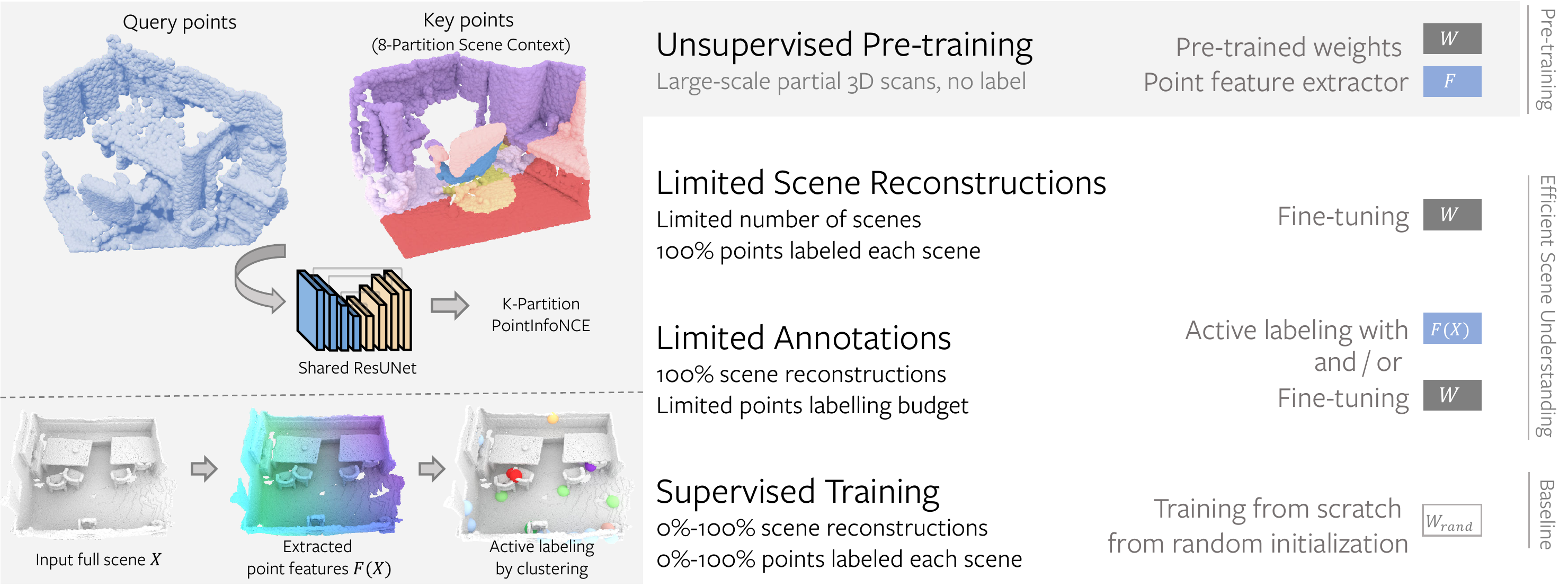}
\end{center}
\vspace{-0.4cm}
\caption{\textbf{Overview of Data-Efficient 3D Scene Understanding.} \textbf{Left}: Unsupervised pre-training with \emph{Contrastive Scene Contexts}. The outputs of pre-training are 1) a pre-trained U-Net {$\mathit{\mathbf{F}}$} (that can be used as an offline feature extractor) and 2) its associated weights {$\mathit{\mathbf{W}}$}. \textbf{Right}: After pre-training, different learning scenarios can be applied for the downstream tasks such as learning with limited scene reconstructions (\emph{LR}) or limited annotations (\emph{LA}). In the case of \emph{LR}, the pre-trained weights {$\mathit{\mathbf{W}}$} are used as network initialization for fine-tuning. In the case of \emph{LA}, all the scene reconstructions can be used but only a limited annotation budget is available, \eg 20 points can be annotated (semantic labels) per scene. Again, {$\mathit{\mathbf{W}}$} can be used as network initialization for fine-tuning; optionally the feature extractor {$\mathit{\mathbf{F}}$} can be used in an active labeling strategy to decide which points to annotate. Baselines are standard supervised learning where models are trained from scratch.}
\label{fig:our_method}
\vspace{-0.3cm}
\end{figure*}

To formally explore data-efficient 3D scene understanding, in this section, we propose two different learning paradigms and relevant benchmarks that are associated with two complementary settings that can occur in real world application scenarios: (1) \emph{limited scene reconstructions (LR)} and (2) \emph{limited annotations (LA)}. The first setting mainly concerns the scenario where the bottleneck of data collection is the \emph{number of scenes} that can be scanned and reconstructed. The second one focuses on the case where in each scene, the budget for labeling is limited (\eg one can only label a small set of points). Since 3D point labeling is human intensive, this represents a practical scenario where a data-efficient learning strategy can greatly reduce the annotation cost. An overview is presented in Figure~\ref{fig:our_method}, and details of individual benchmarks are described below.

 \subsection{Limited Annotations (LA)}
In this benchmark, we explore 3D scene understanding with a limited budget for point cloud annotations. We consider a diverse set of tasks including semantic segmentation, instance segmentation and object detection. Specifically, for instance segmentation and semantic segmentation, the annotation budget is in terms of the \emph{number of points} for labelling. This is practically useful: if an annotator only needs to label the semantic labels for 20 points, it will only require a few minutes to label a full room. Our benchmark considers four different training configurations on ScanNet including using \{20, 50, 100, 200\} labeled points per scene. For object detection, the annotation budget is with respect to the \emph{number of bounding boxes} to label in each scene. Our benchmark considers four different training configurations including \{1, 2, 4, 7\} labeled bounding boxes. Our base dataset is ScanNetV2~\cite{dai2017scannet} which has 1201 scenes for training. We evaluate the model performance on standard ScanNetV2 validation set of 312 scenes that has full labels.

\subsection{Limited Scene Reconstructions (LR)}
For current 3D scene datasets, it is common for annotators to carry commodity depth cameras and record 3D videos at private houses or furniture stores. It might be unrealistic to enter a large number of homes and obtain detailed scanning. In this case, the number of scenes might be the bottleneck and the training has to be done on limited amount of scene reconstructions. We simulate this scenario by random sampling a subset of ScanNetV2 training set. Our benchmark has four configurations \{1\%, 5\%, 10\%, 20\%\} (100\% represents the entire ScanNet train set) for semantic segmentation and instance segmentation; and \{10\%, 20\%, 40\%, 80\%\} for object detection. During test time, evaluation is on all scenes in the validation set. 
\section{Experimental Results}

\label{results}
\vspace{-0.5em}

\begin{figure*}[h]
\begin{center}
\includegraphics[width=0.85\linewidth]{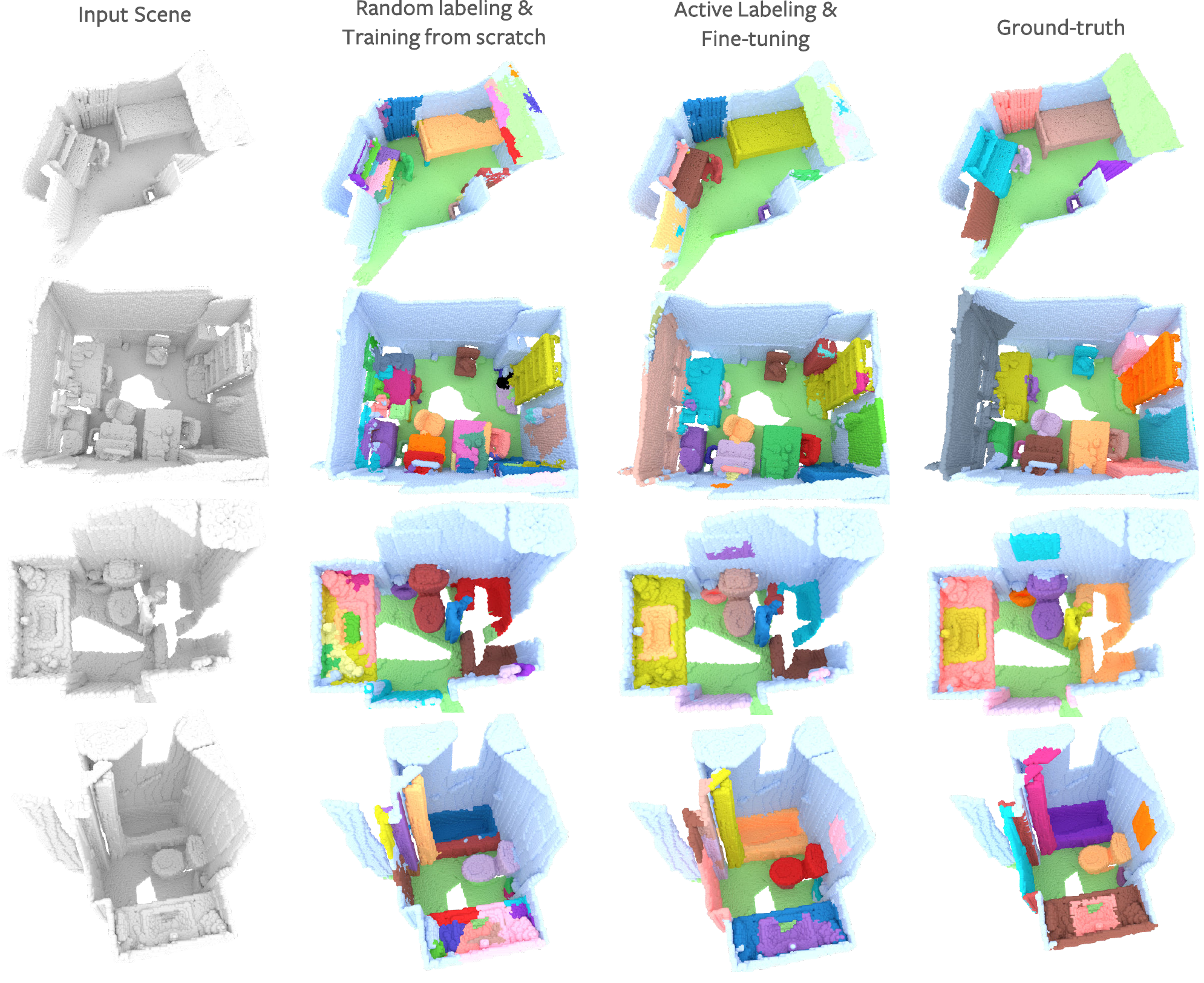}
\end{center}
\vspace{-0.3cm}
   \caption{\textbf{Qualitative Instance Segmentation Results (ScanNet-LA).} With our pre-trained model as initialization for fine-tuning, together with an active labeling process, our approach (trained with 20 labeled points per scene) generates high-quality instance masks. Different color represents instance index only (same instances might not share the same color).}
\label{fig:results}
\vspace{-0.1cm}
\end{figure*}

In this section, we present our experimental results on the data-efficient 3D scene understanding benchmarks: \textbf{ScanNet-LA} with limited annotations and \textbf{ScanNet-LR} with limited scene reconstructions. In both scenarios, we compare our method against the baseline of training from scratch, and report results on semantic/instance segmentation and object detection. We also compare our models with the state-of-the-art method in the last part of the section.

\noindent \textbf{Experiments Setup} For pre-training, we use SGD optimizer with learning rate 0.1 and a batch-size of 32. The learning rate is decreased by a factor of 0.99 every 1000 steps. The model is trained for 60K steps. The fine-tuning experiments on instance segmentation and semantic segmentation are trained with a batch-size of 48 for a total of 10K steps. The initial learning rate is 0.1, with polynomial decay with power 0.9. For all experiments, we use data parallel on 8 NVIDIA V100 GPUs. For object detection experiments, we fine-tune the model with a batch-size of 32 for 180 epochs. The initial learning rate is set to 0.001 and decayed by a factor of 0.1 at epoch 80, 120 and 160. For all the experiments, we use the same Sparse Res-UNet~\cite{xie2020pointcontrast} as the backbone. For both training and testing, the voxel size for Sparse ConvNet is set to 2.0 cm. We use Sparse ConvNet implemented by MinkowskiEngine~\cite{choy20194d}.

\subsection{Limited Annotations}
\vspace{-0.1cm}
As introduced in Section~\ref{benchmark}, the \emph{Limited Annotation (LA)} benchmark covers two different annotation types: \emph{Limited Point Annotations} for semantic and instance segmentation and \emph{Limited Bounding Box Annotations} for detection. The pre-trained network (and its weights) can be used as initialization for fine-tuning, or integrate in an active labeling strategy, which we describe below. 

\noindent\textbf{Active labeling.} Since we focus on the scenario of having limited annotation budget, it is natural to consider an \emph{active learning} strategy during the data annotation process; \ie one can interactively query an annotator to label some data points that can help most for subsequent training. The core idea of our approach is to perform a \textbf{balanced sampling} on the \textbf{feature space}, so that the selected points will be the most representative and exemplary ones in a scene. Our pre-trained network extracts dense features at each point of the to-be-annotated point cloud, by simply performing a forward pass. We then perform k-means clustering in this feature space to obtain $K$ cluster centroids. We select the $K$ centroids as the points to be provided to the annotators for labeling. We also present two baseline strategies including a simple \textbf{random sampling} strategy where $K$ points are randomly selected to be labeled, and a similar \textbf{k-means sampling} strategy on raw (RGB+XYZ) inputs, rather than on the pre-trained features.

We note that although our experiments are simulated based on the already collected ScanNet dataset, our pre-trained feature extractor and the labeling strategy are readily useful in a real-world data annotation pipeline.

\begin{figure*}[t]
\begin{subfigure}{0.3\textwidth}
\includegraphics[scale=0.58]{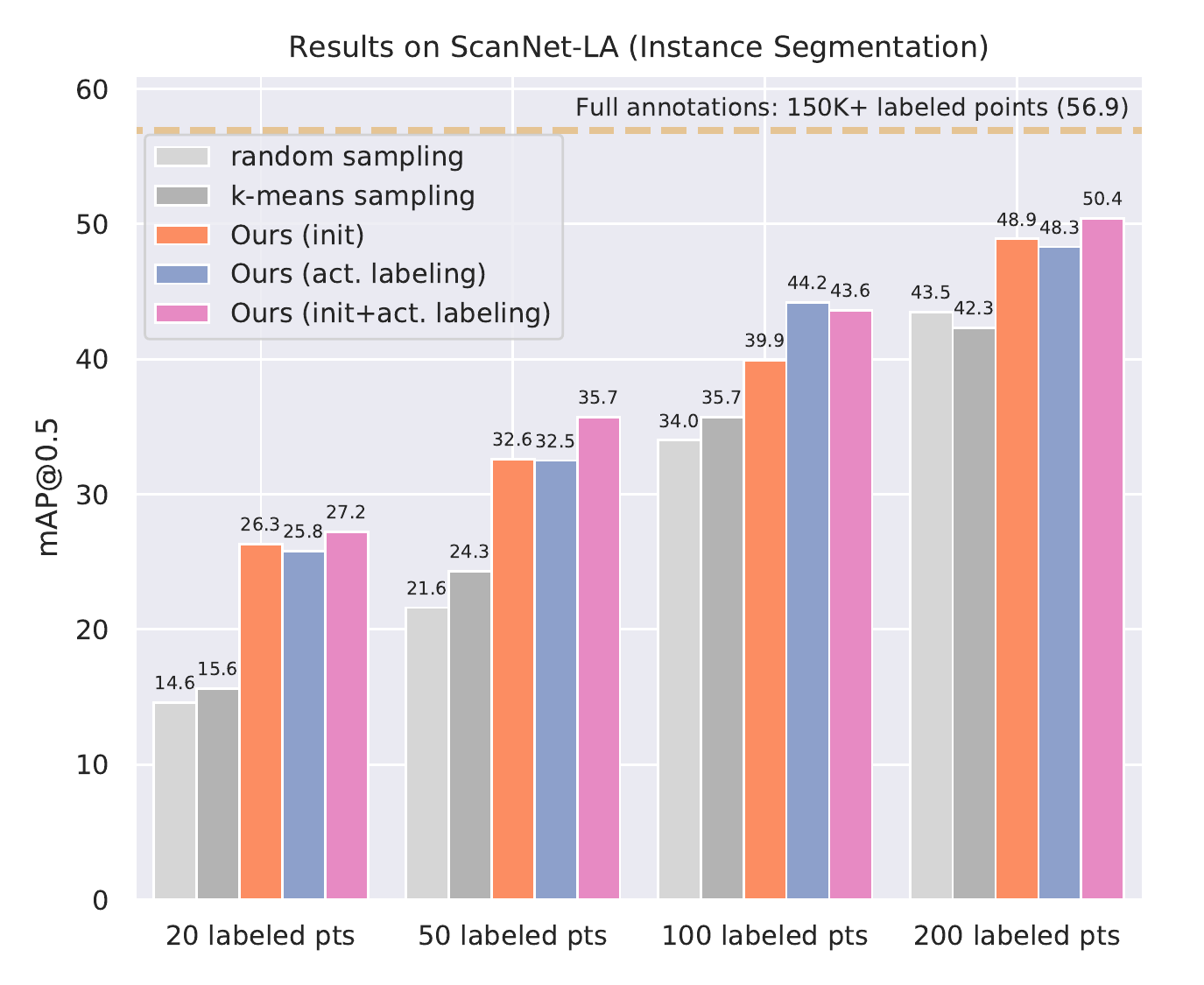}
\end{subfigure} \hspace{0.2\textwidth}
\begin{subfigure}{0.3\textwidth}
\includegraphics[scale=0.58]{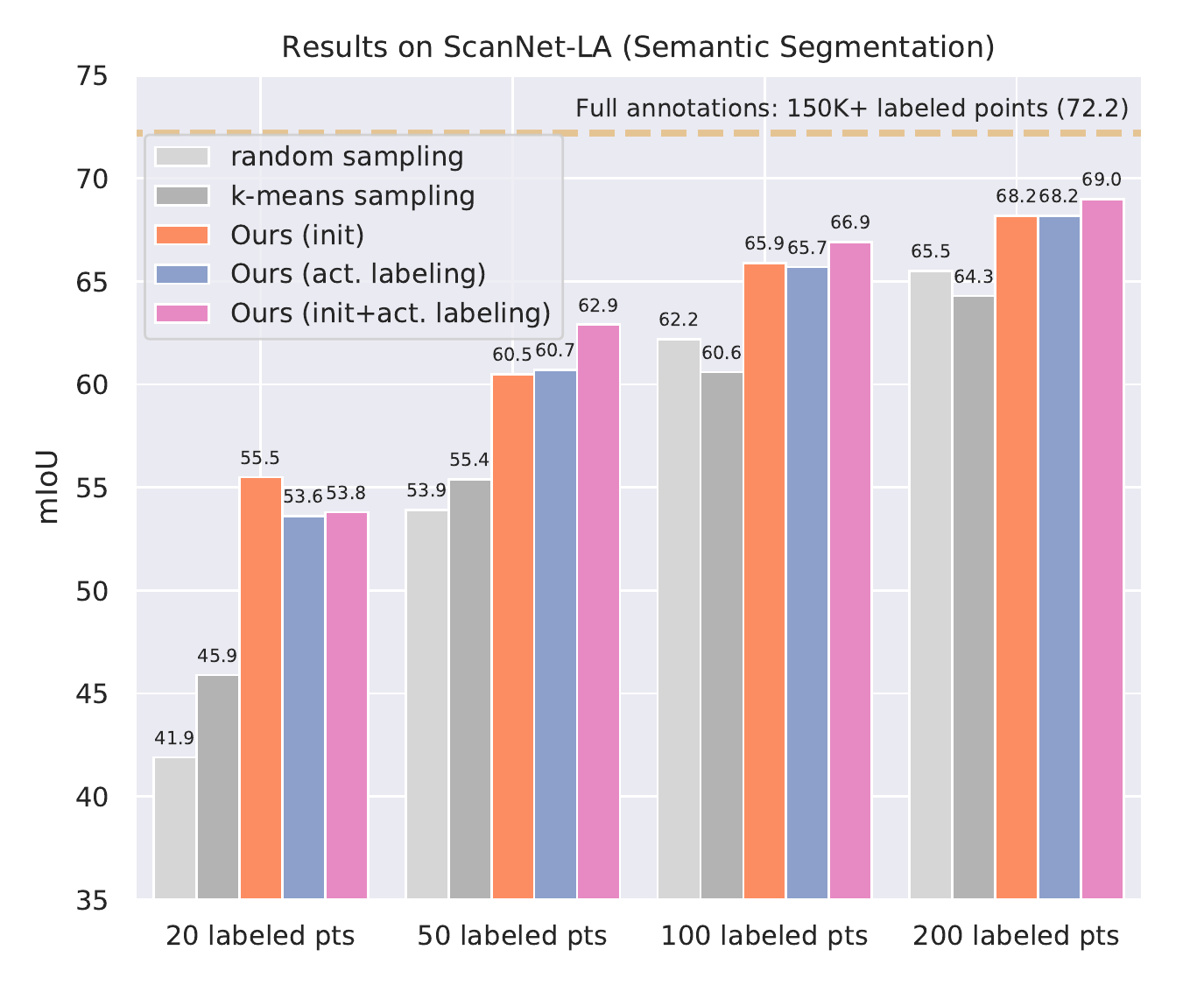}
\end{subfigure}
\vspace{-0.2cm}
\caption{\textbf{3D Instance and Semantic Segmentation with Limited Point Annotations (ScanNet-LA).} Ours (init) denotes the network initialization by our pre-trained model. Ours (act. labeling) denotes the active selection of annotated points by our pre-trained model. Ours (init+act. labeling) denotes using our model as both network initialization and active labeling. We additionally mark the upper bound of using all 150K annotated points (in average) per scene as the dash line.}
\label{fig:bar_allseg}
\vspace{-0.2cm}
\end{figure*}

\noindent\textbf{Results.} In Figure~\ref{fig:bar_allseg} we show that compared to the naive from-scratch baselines, our proposed pre-training framework can lead to much improved performance.  
It is interesting to see that, for both semantic segmentation and instance segmentation, even \emph{without} fine-tuning, the \emph{active labeling} strategy alone provides point labels that make the trained model perform significantly better, compared to \emph{random sampling} or \emph{k-means sampling} baseline strategies, yielding a $>$10\% absolute improvement in terms of mAP@0.5 and mIoU when the training data has only 20 point labels.

\begin{table}[h]
  \centering
  \small
  \begin{tabular}{c|c|c}
  \specialrule{1.1pt}{0.1pt}{0pt}
   No. of Boxes & VoteNet (scratch) & VoteNet (ours)  \\
  \hline
  \rowcolor{Gray} all & 35.4 & \textbf{39.3}~\tiny{\textcolor{darkgreen}{(+3.9)}}\\
   \hline
   1 & 9.1 & \textbf{10.9}~\tiny{\textcolor{darkgreen}{(+1.8)}} \\
   2 & 15.9 & \textbf{18.5}~\tiny{\textcolor{darkgreen}{(+2.6)}} \\
   4 & 22.5 & \textbf{26.1}~\tiny{\textcolor{darkgreen}{(+3.6)}} \\
   7 & 26.5 & \textbf{30.4}~\tiny{\textcolor{darkgreen}{(+3.9)}} \\
 
  \specialrule{1.1pt}{0.1pt}{0pt}
  \end{tabular}
  \vspace{-0.2cm}
  \caption{\textbf{Object detection results using Limited Bounding Box Annotations on ScanNet.} The metric is mAP@0.5. ``Ours" denotes the fine-tuning results with our pre-trained model. We list the upper-bound performance using all annotated bounding boxes (in average about 13 bounding boxes per scene) as a reference in the first row.}
\label{tab:det_less_ann}
\vspace{-0.4cm}
\end{table}

The fact that \emph{active labeling} strategy performs on par with the more common pre-training and fine-tuning paradigm, suggests that finding exemplary points to label is crucial for data-efficient learning. Of course, in real applications both active labeling and fine-tuning can be used jointly, and we indeed observe a further (though admittedly smaller) boost in performance by 1) active sampling points to label and then 2) fine-tuning with the pre-trained weights. 

Overall, with the help of our \OURS{} pre-training, even using around $0.1\%$ of point labels (\eg 200 labeled points out of 150K total points per scene), we are still able to achieve $50.4\%$ mAP@0.5 for instance segmentation, and $69.0\%$ mIoU for semantic segmentation. This indicates a recovery of $89\%$ and $96\%$ of baseline performance that uses $100\%$ of the annotations. We show additional qualitative comparison in Figure~\ref{fig:results}.

\noindent\textbf{Limited Bounding Box Annotations.}  For object detection, we use VoteNet~\cite{voteNet} as the detector framework; follwoing~\cite{xie2020pointcontrast}, we replace PointNet~\cite{qi2017pointnet} with our Sparse Res-UNet. For this part, we do not use any active labeling strategy as the labeling cost for bounding boxes are much smaller. We random sample \{1, 2, 4, 7\} bounding boxes per scene and train the detector. In Table~\ref{tab:det_less_ann}, we observe that our pre-training also consistently improves over the baseline VoteNet, and the performance gap does not diminish when more box annotations are available.

\subsection{Limited Scene Reconstructions}
\vspace{-0.1cm}
In this section, we report the experimental results for another scenario of data-efficient 3D scene understanding, when there is a shortage of scene reconstructions. For instance segmentation and semantic segmentation tasks, we random sample subsets of ScanNet scenes of different sizes. We sample \{1\%, 5\%, 10\%, 20\%\} of the entire 1201 scenes in the training set (which corresponds to 12, 60, 120, and 240 scenes, respectively). For object detection, we find it very difficult to train the detector when the scenes are too scarce. Thus we sample \{10\%, 20\%, 40\%, 80\%\} subsets. For each configuration, we randomly sample 3 subsets and report the averaged results to reduce variance. We also use the official ScanNetV2 validation set for evaluation.

Network fine-tuned with our pre-trained model again shows a clear gap compared to the training from scratch baseline (Table~\ref{tab:allseg_less_scenes}). We achieve competitive results (50.6\% mAP@0.5 for instance segmentation and 64.6\% mIoU for semantic segmentation) using only 20\% of the total scenes.

\begin{table}[]
\centering
\begin{tabular}{c|cc|cc}
\specialrule{1.1pt}{0.1pt}{0pt}
                                & \multicolumn{2}{c|}{Instance Seg.} & \multicolumn{2}{c}{Semantic Seg.} \\
\multirow{-2}{*}{Data Pct.} & Scratch         & Ours        & Scratch         & Ours        \\
\hline
\rowcolor{Gray} 100\%                              & 56.9            & \textbf{59.4}~\tiny{\textcolor{darkgreen}{(+2.5)}}        & 72.2            & \textbf{73.8}~\tiny{\textcolor{darkgreen}{(+1.6)}}        \\
\hline
1\%                                & 9.9             & \textbf{13.2}~\tiny{\textcolor{darkgreen}{(+3.3)}}        & 26.0            & \textbf{28.9}~\tiny{\textcolor{darkgreen}{(+2.9)}}        \\
5\%                                & 31.9            & \textbf{36.3}~\tiny{\textcolor{darkgreen}{(+4.4)}}        & 47.8            & \textbf{49.8}~\tiny{\textcolor{darkgreen}{(+2.0)}}        \\
10\%                               & 42.7            & \textbf{44.9}~\tiny{\textcolor{darkgreen}{(+2.2)}}        & 56.7            & \textbf{59.4}~\tiny{\textcolor{darkgreen}{(+2.7)}}        \\
20\%                               & 48.1            & \textbf{50.6}~\tiny{\textcolor{darkgreen}{(+2.5)}}        & 62.9            & \textbf{64.6}~\tiny{\textcolor{darkgreen}{(+1.7)}} \\
\specialrule{1.1pt}{0.1pt}{0pt}
\end{tabular}
\vspace{-0.2cm}
\caption{\textbf{3D semantic and instance segmentation results with Limited Scene Reconstructions (ScanNet-LR).} Metric is mAP@0.5 for instance segmentation and mIoU for semantic segmentation. ``Scratch'' denotes the training from scratch baseline, and ``Ours'' denotes the fine-tuning results using our pre-trained weights. Results using 100\% of the data during training  are listed in the first row.}
\label{tab:allseg_less_scenes}
\vspace{-0.3cm}
\end{table}

Similar behavior can be observed on the object detection task on ScanNet, and the difference between with and without our pre-training is more pronounced in Table~\ref{tab:det_less_scenes}: the detector can barely produce any meaningful results when the data is scarce (\eg 10\% or 20\%) and trained from scratch. However, fine-tuning with our pre-trained weights, VoteNet can perform significantly better (\eg improve the mAP@0.5 by more than 16\% with 20\% training data).

\begin{table}[h]
  \centering
  \small
  \begin{tabular}{c|c|c}
  \specialrule{1.1pt}{0.1pt}{0pt}
   Data Pct. & VoteNet (scratch)  & VoteNet (ours) \\
   \hline
 \rowcolor{Gray} 100\%  & 35.4 & \textbf{39.3}~\tiny{\textcolor{darkgreen}{(+3.9)}} \\
 \hline
 10\%  & 0.3  & \textbf{8.6}~\tiny{\textcolor{darkgreen}{(+8.3)}} \\
 20\%  & 4.6  & \textbf{20.9}~\tiny{\textcolor{darkgreen}{(+16.3)}} \\
 40\%  & 22.0 & \textbf{29.2}~\tiny{\textcolor{darkgreen}{(+7.2)}} \\
 80\%  & 33.7 & \textbf{36.7}~\tiny{\textcolor{darkgreen}{(+3.0)}} \\
\specialrule{1.1pt}{0.1pt}{0pt}
  \end{tabular}
   \vspace{-0.2cm}
  \caption{\textbf{Object detection results with Limited Scene Reconstructions on ScanNet.} Metric is mAP@0.5. We show constantly improved results over training from scratch, especially so when 10\% or 20\% of the data are available. Results using all scenes are listed in the first row.}
\label{tab:det_less_scenes}
\vspace{-0.3cm}
\end{table}

\subsection{Additional Comparisons to PointContrast}
\vspace{-0.1cm}
\label{comp_to_pc}
As \OURS{} is closely related to PointContrast~\cite{xie2020pointcontrast}, we provide additional results in this section, including comparisons on the data-efficient ScanNet benchmarks (Table~\ref{tab:data_efficient}) as well as on other datasets and benchmarks (Table~\ref{tab:dataset_summary}). Our pre-training method outperforms~\cite{xie2020pointcontrast} in almost every benchmark setting, sometimes by a big margin. These results further render the importance of integrating scene contexts in contrastive learning. Notably, our pre-training method on S3DIS achieves 72.2\% mIoU which outperforms, for the first time, the \emph{supervised} pre-training result reported in~\cite{xie2020pointcontrast}.

\begin{table}[h]
  \centering
  \small
  \begin{tabular}{c|c|cc|c}
  \specialrule{1.1pt}{0.1pt}{0pt}
  Settings & Task (Metric) & SC & PC~\cite{xie2020pointcontrast} & Ours \\
  \hline
 LA (200 points) & ins (\scriptsize{mAP@0.5}) &  43.5 & 44.5~\tiny{\textcolor{gray}{(+1.0)}}& \textbf{48.9}~\tiny{\textcolor{darkgreen}{(+5.4)}} \\
 LA (200 points) & sem (\scriptsize{mIoU})    &  65.5 & 67.8~\tiny{\textcolor{gray}{(+2.3)}}& \textbf{68.2}~\tiny{\textcolor{darkgreen}{(+2.7)}} \\
 LA (7 bboxes  ) & det (\scriptsize{mAP@0.5}) &  26.5 & 28.9~\tiny{\textcolor{gray}{(+2.4)}}& \textbf{30.4}~\tiny{\textcolor{darkgreen}{(+3.9)}} \\
 LR (240 scenes) & ins (\scriptsize{mAP@0.5}) &  48.1 & 48.4~\tiny{\textcolor{gray}{(+0.3)}}& \textbf{50.6}~\tiny{\textcolor{darkgreen}{(+2.5)}} \\
 LR (240 scenes) & sem (\scriptsize{mIoU})    &  62.9 & 63.0~\tiny{\textcolor{gray}{(+0.1)}}& \textbf{64.6}~\tiny{\textcolor{darkgreen}{(+1.7)}} \\
 LR (960 scenes) & det (\scriptsize{mAP@0.5}) &  33.7 & 36.3~\tiny{\textcolor{gray}{(+2.6)}}& \textbf{36.7}~\tiny{\textcolor{darkgreen}{(+3.0)}} \\
 \specialrule{1.1pt}{0.1pt}{0pt}
  \end{tabular}
   \vspace{-0.2cm}
\caption{\textbf{Comparisons to PointContrast for data-efficient 3D scene understanding on ScanNet.} We compare our method with PointContrast (PC) and training from scratch (SC) in various tasks. Our method constantly achieves better results in both Limited Point Annotations (LA) and Limited Scene Reconstructions (LR) scenarios.}
\label{tab:data_efficient}
\vspace{-0.3cm}
\end{table}

\begin{table}[h]
  \centering
  \small
  \begin{tabular}{c|c|cc|c}
  \specialrule{1.1pt}{0.1pt}{0pt}
  Datasets & Task (Metric) & SC & PC~\cite{xie2020pointcontrast} & Ours \\
  \hline
S3DIS     & ins (\scriptsize{mAP@0.5})   & 59.3 & 60.5~\tiny{\textcolor{gray}{(+1.2)}} & \textbf{63.4}~\tiny{\textcolor{darkgreen}{(+4.1)}}\\
S3DIS     & sem (\scriptsize{mIoU})      & 68.2 & 70.3~\tiny{\textcolor{gray}{(+2.1)}} & \textbf{72.2}~\tiny{\textcolor{darkgreen}{(+4.0)}} \\
SUN RGB-D & det (\scriptsize{mAP@0.5} )  & 31.7 & 34.8~\tiny{\textcolor{gray}{(+3.1)}} & \textbf{36.4}~\tiny{\textcolor{darkgreen}{(+4.7)}} \\
ScanNet   & ins (\scriptsize{mAP@0.5})   & 56.9 & 58.0~\tiny{\textcolor{gray}{(+1.1)}} & \textbf{59.4}~\tiny{\textcolor{darkgreen}{(+2.5)}}\\
ScanNet   & sem (\scriptsize{mIou})   & 72.2 & \textbf{74.1}~\tiny{\textcolor{darkgreen}{(+1.9)}} & 73.8~\tiny{\textcolor{gray}{(+1.6)}} \\
ScanNet   & det (\scriptsize{mAP@0.5})   & 35.4 & 38.0~\tiny{\textcolor{gray}{(+2.6)}} & \textbf{39.3}~\tiny{\textcolor{darkgreen}{(+3.9)}} \\
 \specialrule{1.1pt}{0.1pt}{0pt}
  \end{tabular}
   \vspace{-0.2cm}
\caption{\textbf{Downstream fine-tuning results on other benchmarks.} Contrastive Scene Contexts (Ours) achieve better or on par results compared to PointContrast (PC)~\cite{xie2020pointcontrast} on instance segmentation (ins), semantic segmentation (sem) and object detection (det) across multiple datasets.}
\label{tab:dataset_summary}
\vspace{-0.3cm}
\end{table}

\begin{figure}[!htp]
\small
\centering
\includegraphics[width=0.95\linewidth]{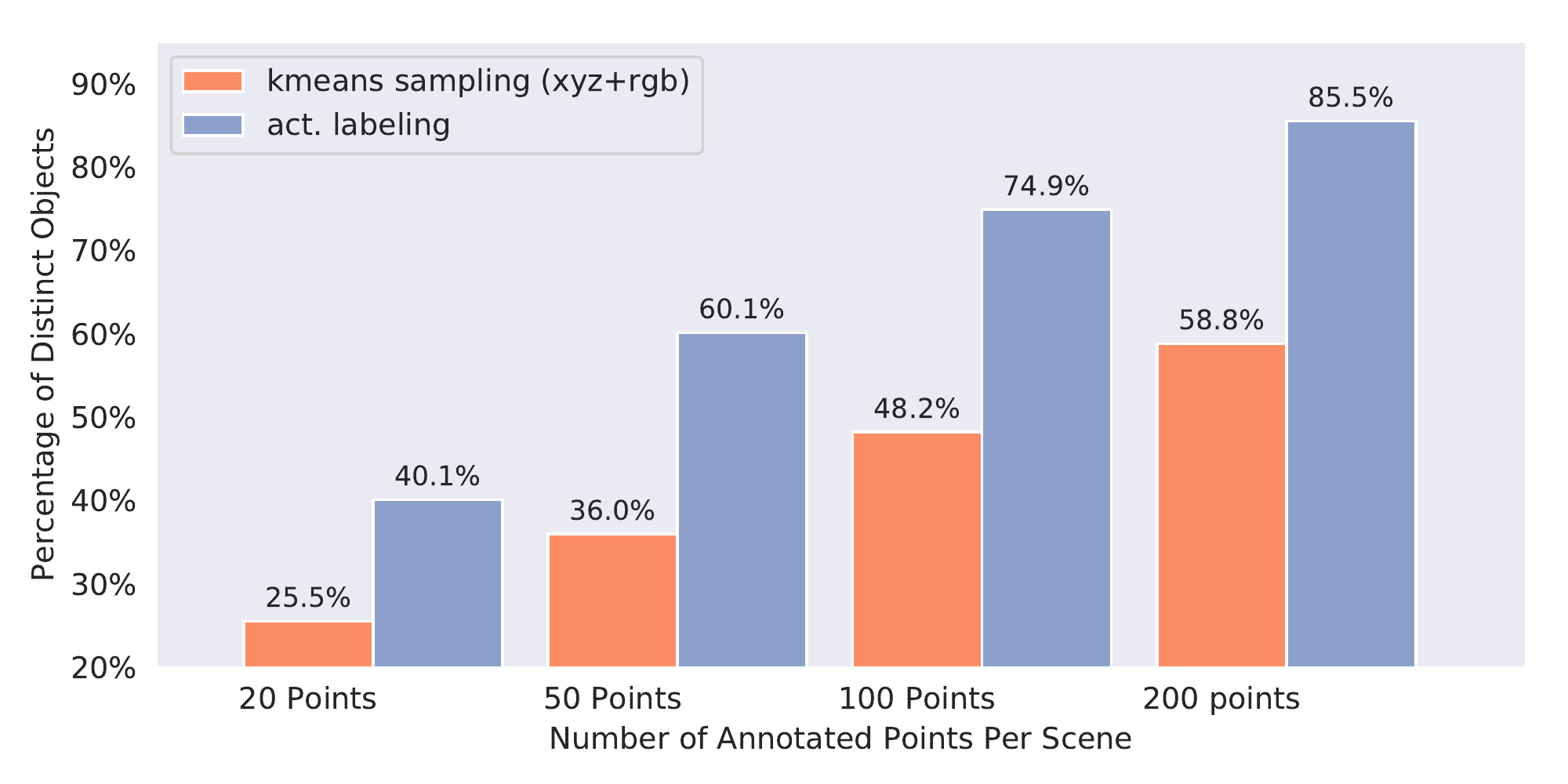}
\includegraphics[width=0.95\linewidth]{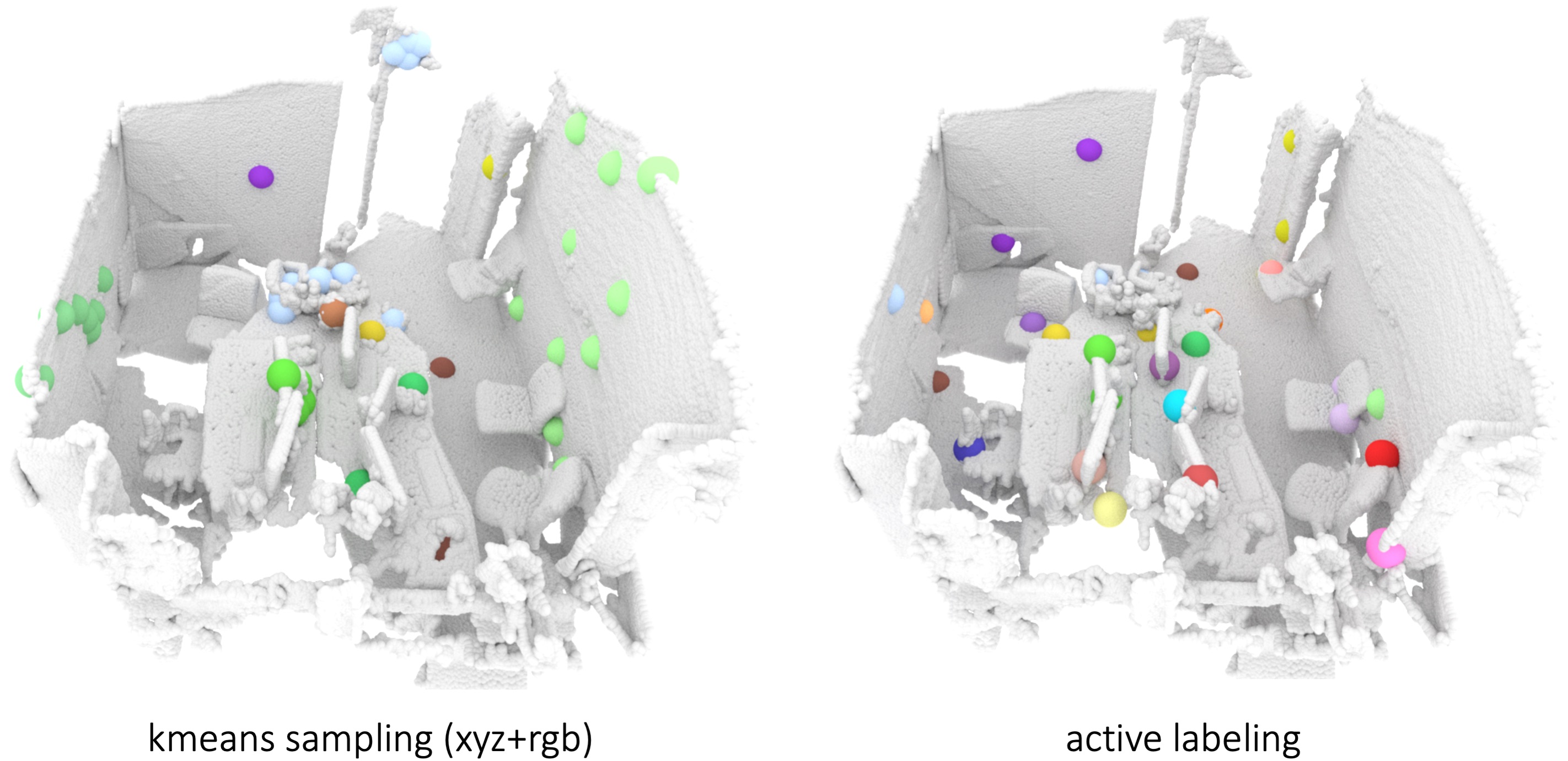} 
\caption{\textbf{Top}: object coverage percentage---more distinct objects are covered with active labeling; \textbf{Bottom}: Visualization of sampled points in a cluttered scene.}
\vspace{-2.0em}
\label{fig:act_vis}
\end{figure}

\subsection{Analysis on Active Labeling: Cluttered Scenes}
To better explain our active labeling strategy and show that it can work in scenes with heavy occlusion and clutter, we filter out a ScanNet subset of 200 cluttered scenes that has multiple objects per one square meter area. Compared to naive k-means sampling, active labeling performs even better on cluttered scenes. In Figure~\ref{fig:act_vis}, we visualize a cluttered scene and sampled points (bottom); we also show quantitatively (top) our strategy covers more distinct objects and thus has a balancing effect.

\section{Conclusion}
In this work, we focus on data-efficient 3D scene understanding through a novel unsupervised pre-training algorithm that integrates the scene contexts in the contrastive learning framework. We show the possibility of using extremely few data or annotations to achieve competitive performance leveraging representation learning. Our results and findings are very encouraging and can potentially open up new opportunities in 3D (interactive) data collection, unsupervised 3D representation learning, and large-scale 3D scene understanding.

{\small \noindent \textbf{Acknowledgments} Work done during Ji's internship at FAIR. Matthias Nie{\ss}ner was supported by ERC Starting Grant \emph{Scan2CAD (804724)}. The authors would like to thank Norman M\"uller, Manuel Dahnert, Yawar Siddiqui and Angela Dai and anonymous reviewers for their constructive feedback.}%

{\small
\bibliographystyle{ieee_fullname}
\bibliography{egbib}
}%

\clearpage
\noindent \textbf{\Large Appendix} \\ 
\begin{appendix}
In this supplemental document, we describe the details of our implementation in Section~\ref{sec:details}. We show more visualizations of our models on semantic segmentation and object detection tasks with extremely scarce data for training in Section~\ref{sec:vis}. Detailed per-category results on data-efficient benchmark as well as on full data are showed in Section~\ref{sec:classes}.

\section{Implementation Details}\label{sec:details}
\paragraph{Data Preprocessing.} Following~\cite{xie2020pointcontrast}, we subsample the partial frames by every 25 frames. We find pairs of frames within each scene by computing their overlaps. In detail, every single frame is transformed to world coordinates. We iterate every pair of frames to calculate how many points are overlapped by 2.5cm threshold. For example, for each point in frame A, if we can find another point in frame B within 2.5cm in the transformed coordinate system (world), then those 2 points are stored as a correspondence pair. When 2 frames have at least 30\% overlaps of points, those 2 frames are saved for training. We save and use both the xyz coordinates and rgb color for pre-training.\\

\vspace{-0.5cm}
\paragraph{PointInfoNCE Loss.} Here we explain the details of the  PointInfoNCE loss (Equation 3 in the main paper).

 {
\begin{align*}
    &\mathcal{L}_p = -\sum_{(i,j) \in M}\log\frac{\exp(\mathbf{f}^1_i\cdot\mathbf{f}^2_j/\tau)}{\sum_{(\cdot, k)\in M,k\in par_{p}(i)}\exp(\mathbf{f}^1_i\cdot\mathbf{f}^2_k/\tau)}
\end{align*}
 }%
 $M$ denotes the set of all the corresponding matches from two frames. Denote the point features from two frames $\textbf{f}^1$ and $\textbf{f}^2$ respectively. In this formulation, we use the points that have at least one match as negative, and non-matched points are discarded. For a matched pair $(i,j) \in M$, point feature $\textbf{f}^1_i$ serves as the query and $\textbf{f}^2_j$ serves as the positive key. Point feature $\textbf{f}^2_k$ where $\exists (\cdot, k) \in M$, $k\in par_{p}(i)$ and $k \neq j$ are used as the set of negative keys. In practice, we sample a subset of matched pairs from $M$ for training.

\paragraph{Active Labelling.} We first use our pre-trained network to make a forward pass on all the voxels of each scene in the training data, and save the 96-dim penultimate layer features at each voxel. Then we back-project the features at each voxel to the raw point cloud using nearest neighbour search. We run a k-means clustering algorithm on the features and xyz coordinates of the point cloud on each scene to get k centroids, where k is the number of points we propose to annotator to label. We run k-means for 50 iterations.

\vspace{-0.2cm}

\paragraph{Clustering Algorithm in Instance Segmentation.} We adapt the code of breadth first search from PointGroup~\cite{jiang2020pointgroup}. Clustering only happens in the test time. In the test time, we cluster on points that are shifted by learned directional and distance vectors. Directional and distance vectors are learned by voting-center loss in the training time. We use 3cm-ball as threshold for every point to search its neighbouring points at each iteration. Within the ball, the points are grouped into one instance when they have the same semantic label. We don't use the ScoreNet proposed in PointGroup, so that we don't have additional network for training. We simply average the scores of semantic prediction of the points belonging to the same instance.


\begin{figure*}[h!]
\begin{center}
\includegraphics[width=1.0\linewidth]{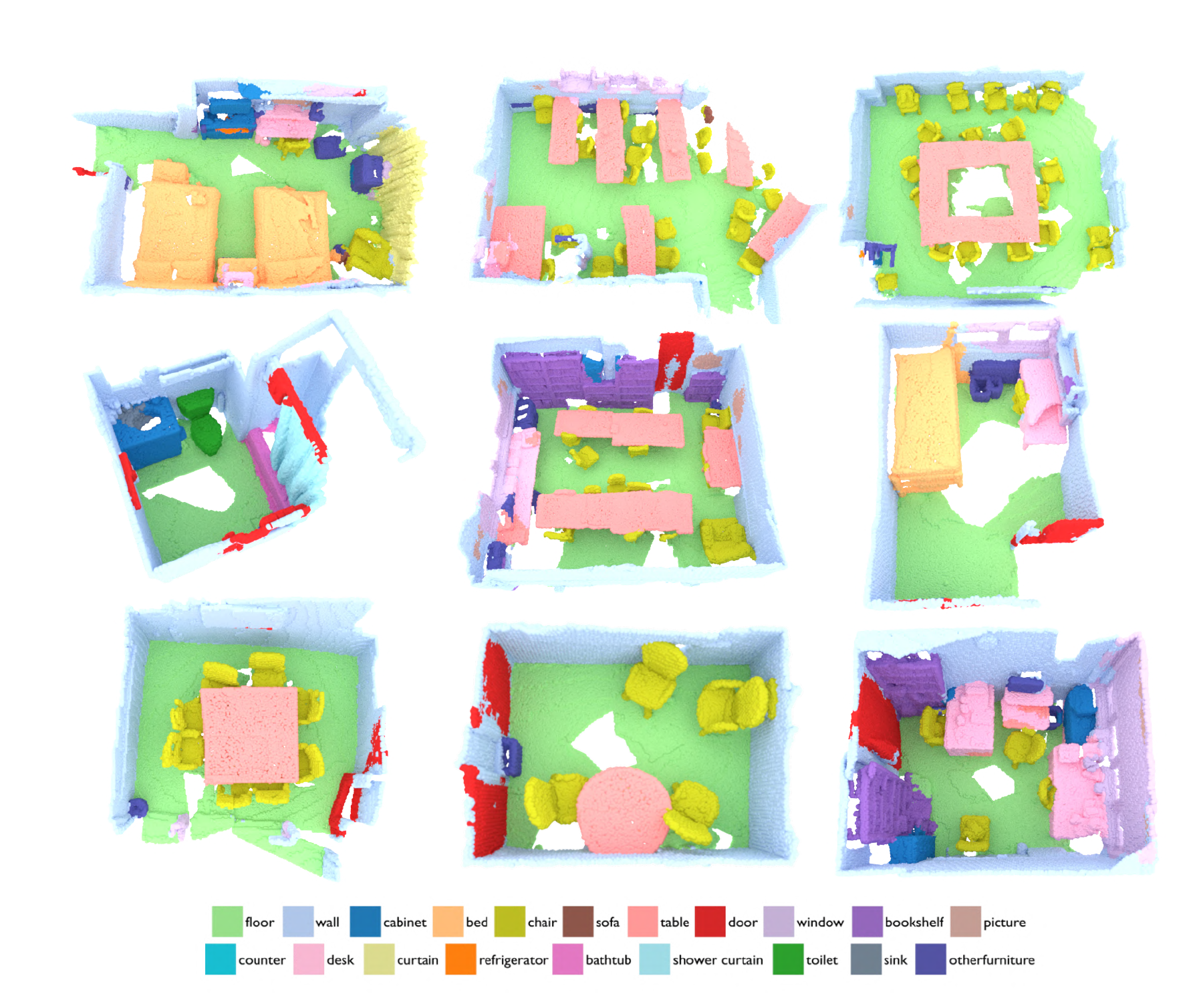}
\end{center}
   \caption{\textbf{Semantic Segmentation Results (ScanNet-LA).} With our pre-trained model as initialization for fine-tuning, together with an active labeling process, our approach generates high-quality semantic segmentation predictions. Here our model is fine-tuned with 20 labeled points per scene.}
\label{fig:sem}
\end{figure*}

\begin{table*}[h!]
    \centering
     \resizebox{\textwidth}{!}{
     \begin{tabular}{l|cccccccccccccccccccc||c}\specialrule{1.3pt}{0.0pt}{0.1pt}
      & cab & bed & chair & sofa & tabl & door & wind & bkshf & pic & cntr & desk & curt & fridg & showr & toil & sink & bath & ofurn & \textbf{avg}\\ \hline
     Scratch& 31.8&72.4&56.0&52.7&\textbf{55.9}&36.6&25.3&47.6&14.7&11.3&10.1&\textbf{36.4}&34.5&57.5&90.0&33.7&\textbf{80.3}&35.8&43.5\\
     PointContrast&39.2&71.2&\textbf{63.1}&\textbf{71.4}&48.4&36.9&20.5&45.2&18.2&8.1&13.9&32.4&31.5&\textbf{64.1}&97.0&42.3&54.9&\textbf{40.1}&44.5\\ \hline
    Ours&\textbf{43.7}&\textbf{75.2}&62.9&65.7&50.5&\textbf{43.4}&\textbf{27.4}&\textbf{52.9}&\textbf{26.9}&\textbf{19.7}&\textbf{14.4}&34.4&\textbf{39.9}&61.9&\textbf{97.4}&\textbf{49.4}&75.3&39.0&\textbf{48.9} \\ \hline
    \end{tabular}
    }
    \vspace{-0.2cm}
    \caption{\textbf{Instance Segmentation with Limited Point Annotations (ScanNet-LA)}. We use mAP@0.5 as metric and demonstrate per-category performance over 18 classes on data-efficient benchmark (200 labelled points for training per scene). }
    \label{tab:la_ins}
     \vspace{-0.4cm}
\end{table*}

\begin{table*}[h!]
    \centering
     \resizebox{\textwidth}{!}{
     \begin{tabular}{l|cccccccccccccccccccc||c}\specialrule{1.3pt}{0.0pt}{0.1pt}
      & wall &floor & cab & bed & chair & sofa & tabl & door & wind & bkshf & pic & cntr & desk & curt & fridg & showr & toil & sink & bath & ofurn & \textbf{avg}\\ \hline
     Scratch& 81.6&\textbf{96.1}&57.5&79.5&88.1&82.2&67.1&55.9&54.4&76.3&24.3&59.9&52.9&67.9&39.8&55.9&86.9&58.2&82.4&42.1&65.5\\
     PointContrast&83.0&96.0&\textbf{61.1}&\textbf{79.5}&89.5&81.9&\textbf{71.6}&57.1&\textbf{57.0}&73.0&22.6&62.0&\textbf{58.8}&\textbf{69.1}&44.4&63.6&91.5&\textbf{59.4}&\textbf{85.2}&48.5&67.8\\ \hline
    Ours&\textbf{84.0}&95.9&60.2&79.0&\textbf{89.5}&\textbf{83.8}&69.6&\textbf{60.2}&56.7&\textbf{80.6}&\textbf{26.1}&\textbf{63.9}&55.6&63.5&\textbf{45.1}&\textbf{63.7}&\textbf{91.9}&56.9&84.7&\textbf{52.6}&\textbf{68.2} \\ \hline
    \end{tabular}
    }
    \vspace{-0.2cm}
    \caption{\textbf{Semantic Segmentation with Limited Point Annotations (ScanNet-LA)}. We evaluate mean IoU over 20 classes on data-efficient benchmark (200 labelled points per scene for training).}
    \label{tab:la_sem}
\end{table*}


\begin{table*}[h!]
    \centering
     \resizebox{\textwidth}{!}{
     \begin{tabular}{l|cccccccccccc||c}\specialrule{1.3pt}{0.0pt}{0.1pt}
      & ceiling &floor & wall & beam & column & window & door &chair& table & bookcase & sofa & board & \textbf{avg}\\ \hline
     Scratch& 46.8 & 89.5 & 72.5 & 0.0 & \textbf{38.2} & 72.5&89.5&88.0&39.3&34.7&72.7&85.7&59.3\\
     PointContrast&66.0&\textbf{93.0}&73.0&0.0&18.6&72.8&88.3&\textbf{91.4}&42.3&\textbf{29.5}&63.6&88.0&60.5\\ \hline
    Ours&\textbf{74.4}&88.0&\textbf{76.5}&\textbf{0.0}&32.4&\textbf{74.6}&\textbf{96.4}&91.0&\textbf{45.0}&28.8&\textbf{63.6}&\textbf{90.5}& \textbf{63.4} \\ \hline
    \end{tabular}
    }
    \vspace{-0.2cm}
    \caption{\textbf{Instance Segmentation on Stanford Area 5 Test~\cite{armeni_cvpr16}}. We evaluate mAP@0.5 over 12 classes. }
    \label{tab:stanford_ins}
\end{table*}

\begin{table*}[h!]
    \centering
     \resizebox{\textwidth}{!}{
     \begin{tabular}{l|ccccccccccccc||c}\specialrule{1.3pt}{0.0pt}{0.1pt}
      & ceiling &floor & wall & beam & column & window & door &chair& table & bookcase & sofa & board & clutter & \textbf{avg}\\ \hline
     Scratch& 91.5&98.6&84.1&0.0&33.0&56.9&63.9&90.1&81.7&72.5&76.5&77.9&59.6&68.2\\
     PointContrast&93.3&\textbf{98.7}&85.6&\textbf{0.1}&\textbf{45.9}&54.4&67.9&91.6&80.1&\textbf{74.7}&78.2&81.5&62.3&70.3\\ \hline
    Ours&\textbf{95.1}&98.4&\textbf{86.3}&0.0&40.7&\textbf{60.8}&\textbf{85.2}&\textbf{91.8}&\textbf{81.9}&73.9&\textbf{78.9}&\textbf{82.8}&\textbf{62.4}&\textbf{72.2} \\ \hline
    \end{tabular}
    }
    \vspace{-0.2cm}
    \caption{\textbf{Semantic Segmentation on Stanford Area 5 Test~\cite{armeni_cvpr16}}. We evaluate mIoU over 13 classes. }
    \label{tab:stanford_sem}
\end{table*}

\begin{table*}[h!]
    \centering
     \resizebox{\textwidth}{!}{
     \begin{tabular}{l|cccccccccc||c}\specialrule{1.3pt}{0.0pt}{0.1pt}
      & bed & table & sofa & chair& toilet & desk & dresser & night stand & book & bathtub & \textbf{avg}\\ \hline
     Scratch&47.8&19.6&48.1&54.6&60.0&6.3&15.8&27.3&5.4&32.1&31.7\\
     PointContrast~\cite{xie2020pointcontrast}&50.5&19.4&51.8&\textbf{54.9}&57.4&\textbf{7.5}&\textbf{16.2}&37.0&5.9&\textbf{47.6}&34.8\\ \hline
    Ours&\textbf{55.3}&\textbf{20.3}&\textbf{53.8}&53.6&\textbf{65.9}&6.1&15.5&\textbf{38.0}&\textbf{9.1}&46.5&\textbf{36.4} \\ \hline
    \end{tabular}
    }
    \vspace{-0.2cm}
    \caption{\textbf{Object Detection on SUN RGB-D~\cite{song2014sliding}}. We use mAP@0.5 as metric and show per-category AP@0.5 over 10 classes. }
    \label{tab:sunrgbd_det}
\end{table*}

\begin{table*}[h!]
    \centering
     \resizebox{\textwidth}{!}{
     \begin{tabular}{l|cccccccccccccccccc||c}\specialrule{1.3pt}{0.0pt}{0.1pt}
     & cab & bed & chair & sofa & tabl & door & wind & bkshf & pic & cntr & desk & curt & fridg & showr & toil & sink & bath & ofurn & \textbf{avg}\\ \hline
     Scratch&49.0&70.0&87.4&66.5&71.1&47.4&39.6&53.0&30.8&\textbf{32.8}&30.8&41.7&48.6&60.1&\textbf{99.9}&68.4&75.3&52.4&56.9\\
     PointContrast&49.4&72.1&87.2&\textbf{71.7}&67.0&\textbf{49.0}&40.7&\textbf{57.8}&\textbf{35.6}&24.0&30.2&\textbf{49.9}&\textbf{53.0}&65.2&98.3&61.7&80.5&50.8&58.0\\ \hline
     Ours&\textbf{50.8}&\textbf{74.1}&\textbf{88.7}&61.4&\textbf{67.2}&48.0&\textbf{42.0}&57.0&33.8&32.5&\textbf{42.9}&47.4&49.5&\textbf{68.9}&98.2&\textbf{71.3}&\textbf{80.5}&\textbf{54.7}&\textbf{59.4}\\ \hline
    \end{tabular}
    }
    \vspace{-0.2cm}
    \caption{\textbf{Instance Segmentation on ScanNetV2~\cite{dai2017scannet} Validation Set}. We evaluate the mean average precision with IoU threshold of 0.5 over 18 classes.}
    \label{tab:scannet_ins}
\end{table*}

\begin{table*}[h!]
    \centering
     \resizebox{\textwidth}{!}{
     \begin{tabular}{l|cccccccccccccccccc||c}\specialrule{1.3pt}{0.0pt}{0.1pt}
      &  cab & bed & chair & sofa & tabl & door & wind & bkshf & pic & cntr & desk & curt & fridg & showr & toil & sink & bath & ofurn & \textbf{avg}\\ \hline
     Scratch&9.9&70.5&70.0&60.5&43.4&\textbf{21.8}&10.5&\textbf{33.3}&0.8&15.4&33.3&26.6&\textbf{39.3}&9.7&74.7&23.7&75.8&18.1&35.4\\
     PointContrast&13.1&\textbf{74.7}&\textbf{75.4}&\textbf{61.3}&44.8&19.8&12.9&32.0&0.9&\textbf{21.9}&31.9&\textbf{27.0}&32.6&17.5&\textbf{87.4}&23.2&80.8&\textbf{26.7}&38.0\\ \hline
    Ours&\textbf{15.1}&74.3&71.9&60.2&\textbf{46.4}&21.2&\textbf{15.0}&32.5&\textbf{1.1}&9.4&\textbf{36.6}&21.3&37.3&\textbf{47.5}&84.3&\textbf{26.2}&\textbf{86.8}&21.2&\textbf{39.3} \\ \hline
    \end{tabular}
    }
    \vspace{-0.2cm}
    \caption{\textbf{Object Detection on ScanNetV2 Validation Set}. We use mAP@0.5 as metric and show per-category performance over 18 classes. }
    \label{tab:scannet_det}
\end{table*}

\section{More Visualizations} \label{sec:vis}

We show more visualizations of semantic segmentation and object detection predictions from our model trained with extremely scarce annotations. We show the semantic segmentation on ScanNet validation set with our model trained on 20 labelled points per scene in Figure~\ref{fig:sem}. We also demonstrate the object detection results on ScanNet validation set predicted by our model trained on 1 bounding box annotated per scene in Figure~\ref{fig:det}.\\

\section{Per-Category Results}\label{sec:classes}
In this section, we demonstrate detailed per-category performance as supplement of data-efficient benchmark. Instance segmentation on ScanNet-LA (Limited Scene Annotations, 200 labelled points for training) is showed in Table~\ref{tab:la_ins}; semantic segmentation of per-category performance on ScanNet-LA is showed in Table~\ref{tab:la_sem}; object detection on Limited Bounding Boxes Annotations is showed in Table~\ref{tab:la_det}.

We further show the detailed per-category performance as supplement of Table. 6 in the main paper on full data. Instance segmentation and semantic segmentation results on S3DIS are showed in Table~\ref{tab:stanford_ins} and Table~\ref{tab:stanford_sem}; object detection on SUN-RGBD result is showed in Table~\ref{tab:sunrgbd_det}; instance segmentation and object detection on ScanNet validation set are showed in Table~\ref{tab:scannet_ins} and Table~\ref{tab:scannet_det}.

\section{Different Backbones.}
We use Sparse Residual U-Net (SR-UNet-34, also used in~\cite{choy20194d}) as backbone architecture. 3D-MPA also uses a Sparse Residual U-Net backbone, and the performance gap is due to the additional head modules (e.g., Proposal Consolidation) which is orthogonal to our pre-training method. To show our algorithm is generic and agnostic to the specific backbone, we perform experiments with different backbones, including SR-UNet-18A and PointNet++. Models pre-trained with our method yield significant better results; see Tab.~\ref{tab:backbone}.

\begin{table}[htb]
\small
\centering
\scalebox{0.9}{
\begin{tabular}{c|c|c|c|c}
\specialrule{1.1pt}{0.1pt}{1.2pt}
      & Task & Dataset & Backone & mAP@0.5    \\
\hline
scratch & ins & S3DIS & SR-UNet-18A  & 58.6  \\
ours (pre-trained)   & ins & S3DIS & SR-UNet-18A  &    \textbf{62.8} \\
\hline
scratch & det & ScanNet & PointNet++  & 33.5   \\
ours (pre-trained)   & det & ScanNet & PointNet++  &  \textbf{39.2}  \\
\specialrule{1.1pt}{0.1pt}{2.2pt}
\end{tabular}
}
\caption{Pre-training with different backbones; 100\% of available train data is used; we would expect larger deltas with smaller train set.}
\label{tab:backbone}
\end{table}


\section{ScanNet Benchmark} We report validation results to directly compare with PointContrast which also evaluates on the val set. Additionally, we submitted our model to the ScanNet Benchmark  (test set); see Tab.~\ref{tab:benchmark}. Our method significantly outperforms 3D-MPA, despite not leveraging the special 3D-MPA proposal module.

\begin{table}[htb]
\centering
\small
\begin{tabular}{c|c|c|c}
\specialrule{1.1pt}{0.1pt}{1.2pt}
      & AP & AP@50 & AP@25   \\
\hline
3D-MPA~\cite{engelmann20203d} & 35.5 & 61.1  & 73.7  \\
ours (pre-trained)   & \textbf{40.5} & \textbf{64.8}  & \textbf{79.1}   \\
\specialrule{1.1pt}{0.1pt}{2.2pt}
\end{tabular}
\caption{ScanNet \textbf{test} set: similar to S3DIS, we outperform 3D-MPA.}
\label{tab:benchmark}
\end{table}
\end{appendix}

\end{document}